\documentclass[12pt,british]{article}

\usepackage[T1]{fontenc}
\usepackage[utf8]{inputenc}
\usepackage[letterpaper]{geometry}
\usepackage{verbatim}
\usepackage{float}
\usepackage{calc}
\usepackage{graphicx}

\makeatletter
\usepackage[utf8]{inputenc}

\usepackage{marvosym}
\date{}

\makeatother

\usepackage{babel}
\begin{document}

\title{The Voynich Manuscript is Written in Natural Language: The Pahlavi
Hypothesis}

\author{J. Michael Herrmann\\
University of Edinburgh, School of Informatics\\
10 Crichton St, Edinburgh, EH8 9AB, U.K.}
\maketitle
\begin{abstract}
The late medieval Voynich Manuscript (VM) has resisted decryption
and was considered a meaningless hoax or an unsolvable cipher. Here,
we provide evidence that the VM is written in natural language by
establishing a relation of the Voynich alphabet and the Iranian Pahlavi
script. Many of the Voynich characters are upside-down versions of
their Pahlavi counterparts, which may be an effect of different writing
directions. Other Voynich letters can be explained as ligatures or
departures from Pahlavi with the intent to cope with known problems
due to the stupendous ambiguity of Pahlavi text. While a translation
of the VM text is not attempted here, we can confirm the Voynich-Pahlavi
relation at the character level by the transcription of many words
from the VM illustrations and from parts of the main text. Many of
the transcribed words can be identified as terms from Zoroastrian
cosmology which is in line with the use of Pahlavi script in Zoroastrian
communities from medieval times. 
\end{abstract}

\section{Why is Voynichese difficult? }

All writing systems in the world \cite{Faulmann,Daniels} require
some effort in acquisition and use. While for some groups of languages,
difficulty and differences may be comparatively small \cite{Paulesu},
in others the complexity of the script can appear forbidding for all
but a minority of scribes. Religious observance, for example, may
require the adherents to continue using a script or language that
no longer adapts to its language environment and that may thus tend
to become ambiguous or incomprehensible. In order to retain a unique
pronunciation and, supported by extensive commentaries, continuing
understandability, glyphs (\emph{diacritics}) from were added to letters
to distinguish them, or additional letters (\emph{matres lectionis})
were inserted to represent sounds (such as vowels in the consonant-based
(\emph{abjad}) scripts. However, such additional efforts may not be
considered necessary, if the oral tradition in the community is sufficiently
strong, such that the texts do not have to be extracted from the writing
itself, but are rather remembered while being read. If the Voynich
Manuscript (VM, MS 408 in the Beinecke Rare Book \& Manuscript Library
at Yale University) derives from such a tradition, the difficulty
in reading it may be understandable. 

The Voynich Manuscript (VM, MS 408 in the Beinecke Rare Book \& Manuscript
Library at Yale University) which is written on more than 200 vellum
pages has been dated between 1404 and 1438 (University of Arizona,
2011), but its history is largely unknown until the discovery by the
bookseller Voynich in 1912. Apart from a few cautious attempts, such
as Ref.~\cite{Bax}, so far little progress has been achieved in
deciphering the VM nor even a decision was reached whether the VM
has any meaningful content at all~\cite{Reddy}. 

Our hypothesis that the VM is written in natural language, is to be
evidenced by showing that the script used in the VM is directly related
to Pahlavi, a writing system that was in use for several Iranian languages
from before the current era at least until 900 \cite{GeigerandKuhn2002}.
Pahlavi is a particular case of a language that is notoriously difficult
to read. It was used in medieval scriptures, commentaries, and a few
other texts \cite{Andreas} related to Zoroastrianism, the pre-Islamic
religion of Persia. Over the few centuries of the language evolution,
many Pahlavi letters have coalesced, e.g.~for the phonemes \emph{d},
\emph{g}, \emph{j}, and \emph{y}, only a single letters is retained
in Pahlavi. Moreover, letters are usually joined in Pahlavi script
and can appear thus similar to other letters: E.g., in addition to
its proper meaning, a letter can be indistinguishable from as much
a sixteen different phoneme or letter combinations \cite{JustiBund}.
In some words, corrupted forms of letters %
{} have become a standard that is accepted to various degrees by the
scribes. In addition to Persian words, Pahlavi contains also a large
number of heterograms, i.e.~around a thousand, partially very common
words of Aramaic origin that are meant to be read in Persian (like
the Latin abbreviation \emph{i.e.} is read in English as \emph{that
is}).%
{} Finally, as for many other ancient texts, material decay, language
drift, scribe errors, unfamiliarity with the original cultural context,
and, possibly, the need of the writers to hide the content from contemporary
hostility, also contribute to the difficulty of reading the text.

Concerning recent work on the VM, statistical approaches \cite{Amancia,Jaskiewicz,Montemurro}
that search for non-random features in data may be bound to fail if
the target is quite random to begin with. The standard Voynich character
set (EVA) \cite{EVA} is not too helpful either, because it is unrelated
to the phonemics, it breaks some of the letters into smaller parts,
and fails to identify ligatures, all of which may further reduce the
strength of the statistical analysis, cf. ~\cite{Rugg2004,Schninner2007,Reddy,Jaskiewicz}.
In addition, the extensive 19th century literature dedicated to religious
writing, see e.g.~\cite{Mueller} was difficult to access until scanned
copies became available online recently, and, finally, it may be construed
that our academic habits thwart the systematic study of matters as
obscure as the VM. 


The Pahlavi hypothesis was proposed informally already in 
2005~\cite{Peterson2005}\footnote{I was not aware of this 
news-group post until I found a 
Twitter comment on the first version of the current paper where  
Ref.~\cite{Peterson2005} was mentioned.}.
The hypothesis is based there on the similarity of the numbers of letters 
(``14 - 17'') in the Voynich and Pahlavi alphabets and on a general 
perception of a topical relation to the Bundahesh and the Denkard.
Also a small sample of words lengths
from a Pahlavi text was included, 
but was not compared to a transliteration of the Voynich text.

The present paper aims at providing evidence for the hypothesis that
the VM is a readable text with an interest in itself. Our approach
consists in establishing a relation between the Voynich and Pahlavi
scripts (see Section 2). It will also become clear that only within
a cooperation among experts in Pahlavi philology, Zoroastrianism,
history of medicine, botany, astronomy and palaeography, the content
of the VM can be revealed.

We will provide evidence for the proposed relation between the two
alphabets by a number of examples from VM illustrations as well as
from its running text (Section 3). Finally, we will draw (in Section
4) some rather speculative conclusions on the context in which the
content of the VM may have originated.

\section{Letters are reverted Pahlavi characters}

Comparing the Voynichese and Pahlavi scripts, we find that many of
the characters are upside-down versions of each other, see Table \ref{tab:Letters-in-brackets}.
This may be due to the different writing direction of the two scripts.
A similar effect that was observed also in the earlier sinistrodextral
Brahmi script \cite{Buehler}, in which also some of the letters appear
as upside-down adoptions from its likely predecessor Aramaic (right
to left). Pahlavi, that ultimately derives also from the Aramaic alphabet,
has retained the dextrosinistral direction, while the VM is written
in the opposite direction.

In this way, six of the about 20 Voynich letters can be explained
directly (\emph{a}, \emph{h}, \emph{s}, \emph{S}, \emph{r}, in our
notation, see Table \ref{tab:Letters-in-brackets}, and \emph{K},
see Table \ref{tab:Ligatures-involving-are}). Two more letters (\emph{d},
\emph{c}) differ from \emph{s} and \emph{S}, respectively, only by
an inverted breve diacritic. In addition, there are three more letters
that obtain by rotation about a different angle (\emph{t},\emph{ y})
or by mirroring (\emph{z}). The similarity of eleven out of the comparatively
small number of letters of the two alphabets can be considered as
a clear indication of a relation between Voynich (V) and Pahlavi (P).
Below we will see that the relation extends also to the phonemic level.
Two letters \emph{o} and \emph{n} that occur frequently in the VM,
differ from their counterpart in the P alphabet. It is tempting to
relate V\emph{ o} to P\emph{ pe}, but we suggest rather an association
to \emph{waw}. This also supported by the frequent use of \emph{o}
as a word separator in the VM. In Pahlavi a vertical bar is used for
this purpose, which is of similar shape as P\emph{ v}, while in the
VM apparently the more distinctive letter \emph{o} has been preferred.
Further analysis of the V text will shown whether \emph{o}, \emph{y}
or \emph{a} also have a grammatical function. Based on the phonetic
content (Sect.~3) of the letters, we assume that, in contrast to
Pahlavi, the nasal alveolar is not part of the spectrum of V \emph{o},
but is represented by V \emph{n}. 

The remaining V letters are the ``capitals'' \emph{B}, \emph{K},
\emph{M}, \emph{P}, or occur only very rarely (see f57v for a number
of other characters, some of which, however, occur nowhere else in
the VM). The shape of the V ``capitals'' may have arisen from a
fusion of the respective Pahlavi characters with a vertical stroke
(P word separator). We do not consider the capitals to be functional
ligatures, though, as they are used also within words or after the
word separator (\emph{o}).

\thispagestyle{empty}
\begin{table}
\begin{centering}
\includegraphics[width=13cm]{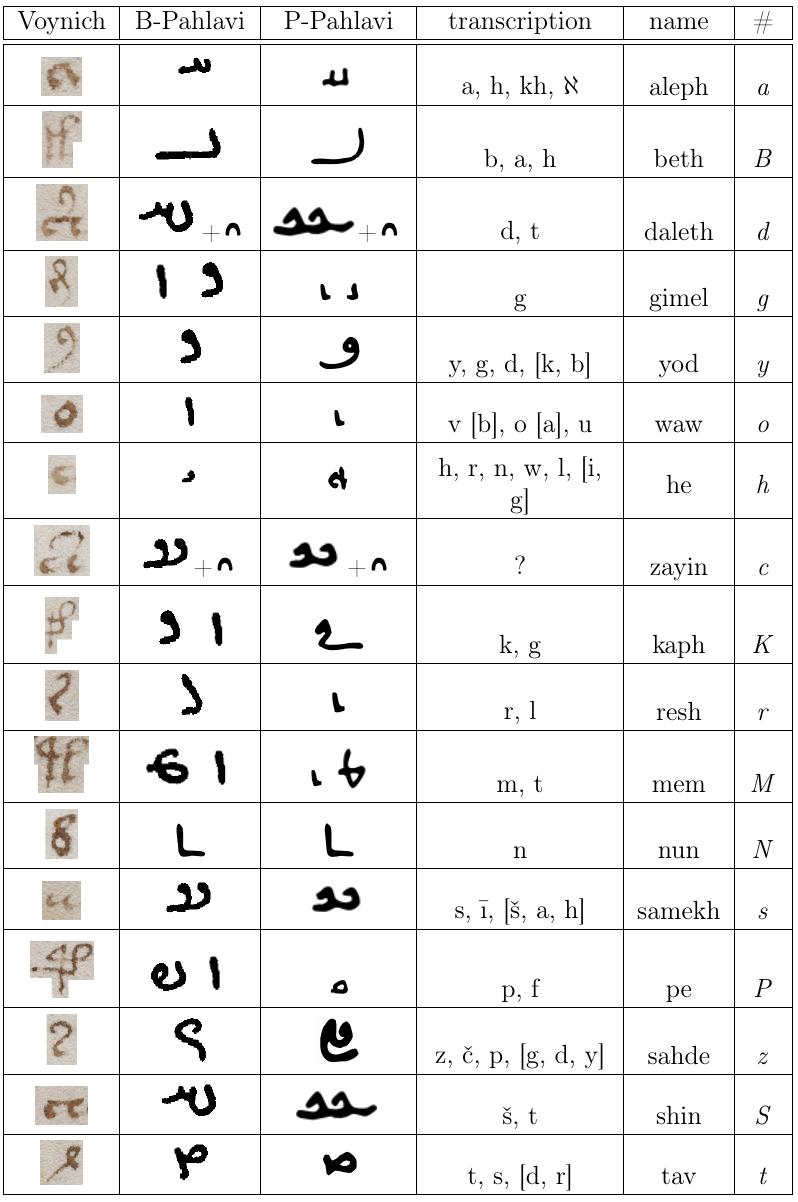}
\par\end{centering}

\caption{Voynich characters and initials together with variants of the corresponding
Pahlavi letters. The last column shows the notation used here. See
Box 1 for comments. \label{tab:Letters-in-brackets}}
\end{table}

\fbox{\begin{minipage}[t]{1\columnwidth}%
Box 1: Comments on Table \ref{tab:Letters-in-brackets}.

The letters are given in the order to the Aramaic alphabet with \emph{resh}
taking the place of phonetically similar \emph{lamed}, and \emph{jod}
is placed near \emph{daleth} and \emph{gimel} with which it is interchangeable
in Pahlavi. Frequently occurring corruptions are given in {[}brackets{]}
\cite{MacKenzie}. Strokes from neighbouring characters are removed
from the Voynich letters.

{[}\emph{a}{]} Appears in B-Pahlavi as a raised character. $\aleph$
represents a glottal stop.\\
{[}\emph{B}{]} We could not find enough evidence for systematic use
of two variants (\emph{B} and \emph{H}) of this character.\\
{[}\emph{g}{]} Occurs usually in final position, elsewhere \emph{y}
is used instead.\\
{[}\emph{o}{]} V \emph{v} resembles Syriac vav (\includegraphics[height=0.3cm]{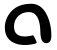}),
Pahlavi \emph{vav} is identical to \emph{resh}\\
{[}\emph{y}{]} the letter represented here is \emph{daleth}. The actual
P-Pahlavi letter \emph{yod} (\includegraphics[height=0.4cm]{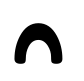})
shows an interesting similarity to the inverted breve diacritic of
V \emph{d} and V \emph{c}. Many words have an otiose \emph{y} ending.\\
{[}\emph{c}{]} This character occurs rarely in the VM, the mere fact
that we did not identify a distinctive character for P \emph{\v c}
does not justify the transliteration of \emph{c} by \emph{\v c.}\\
{[}\emph{z}{]} is often (or easily) confused with \emph{r.}\\
{[}P{]} occurs often at the beginning of paragraphs. It may be an
abbreviation of \emph{pad }for \emph{to}, \emph{at}, \emph{in} or
\emph{on}.%
\end{minipage}}

\begin{table}
\begin{centering}
\includegraphics[width=9cm]{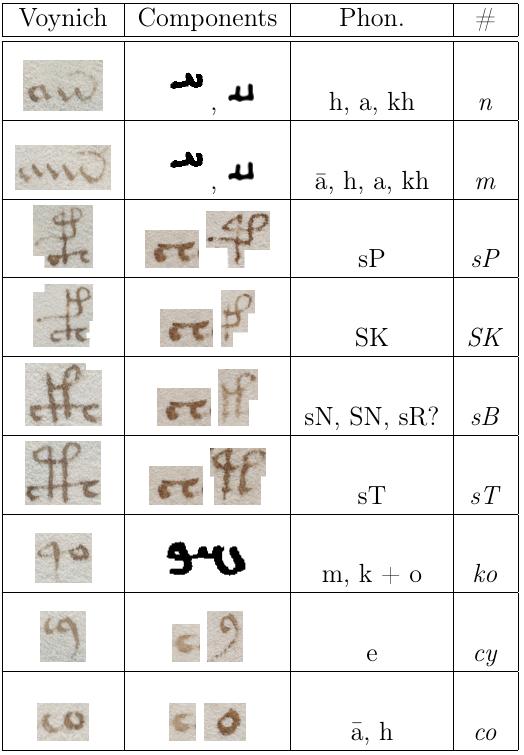}
\par\end{centering}

\caption{Main ligatures and letter combinations from the VM. Only part of the
implied phonemes are given in the third column. The last column refers
to the transliteration in Table \ref{tab:Letters-in-brackets}. The
two or three strokes of \emph{n} or \emph{m} have a similar functions
as \emph{h} in the final or penultimate position. Ligatures involving
the letter V \emph{S} (``table'') represent the succession of two
consonants usually in the beginning of a word. In some cases it is
\emph{s} rather than S that is represented. While \emph{sP} and \emph{sT}
are obvious from the vocabulary, the remaining combinations will have
to be reconsidered. First part of \emph{ko} occurs rarely if ever
alone. This ligature can represent \emph{m}, \emph{q}, \emph{h}, \emph{r},
\emph{mn}, \emph{mv}, \emph{mr}, \emph{m}$\aleph$, etc. The combinations
\emph{cy} and \emph{co} appear to represent single phonemes in some
cases, see Appendix \ref{sec:Sample-text-(f1r)}. All ligatures are
copied from f37r, the componentsin the second column are from Table~\ref{tab:Letters-in-brackets}.
Strokes belonging to neighbouring characters were removed. \label{tab:Ligatures-involving-are}}
\end{table}

\section{Vocabulary relates to Zoroastrian religion}

Voynichese and Pahlavi are not identical. By the introduction of a
number of additional characters, such as to distinguish \emph{d},
\emph{g}, \emph{n}, reading a Voynichese text may have been easier
than a Pahlavi text. It is not clear why the history of the deciphering
of the VM, does not support this claim. Analysing plant and star names,
Bax~\cite{Bax} has suggested a similar reading for some but not
all of the letters. We base our transcription on a larger number of
samples from the manuscript and compare the results with names from
the Zoroastrian cosmological scripture \emph{Bundahesh }\cite{WestBundahesh,JustiBund},
which was composed in the 11th century, and with general vocabulary
\cite{MacKenzie}, such that we arrive at a more complete and more
reliable transcription that is based not only on the similarity of
the letter shapes. The translations given below should not be expected
to do justice to the VM text. They are solely included to provide
evidence for the proposed transcription.

\subsection{Zodiac symbols}

In the appendix, we show two sets of words from the manuscript. The
first (App.~\ref{sec:Zodiac-signs-(f70v1-f73v)}) gives the names
of the zodiac symbols and the corresponding month names both of which
were passed down in the Bundahesh \cite{WestBundahesh,JustiBund}
in paragraphs II, 2 and XXV, 20, respectively. Based on the well known
symbols shown in the centres of f70v1 -- f73v, the identification
with the Pahlavi names is straight-forward, expect for the two pages
f71r and f72r1 which show the same symbols (Aries and Taurus) as f70v1
and f71v, respectively. We cannot answer the question whether the
two repeated signs do in fact represent the missing Capricorn and
Aquarius. Because two words (on f72r2 and f72v2) are unreadable due
to creases, we are left with 18 words that are identifiable to a reasonable
degree of certainty.

\subsection{Plant names }

Ancient plant names are occur in manifold variants and are often ambiguous.
The same seems true for the plant drawings in the VM, where, in some
case, it seems even plausible that the artist followed merely a verbal
description rather than an own view or any original drawings. We can
thus expect only a few characteristics to be identifiable. In addition,
only a few plant names are included in the standard dictionaries (e.g.~\cite{MacKenzie}),
such that most of the VM plant depictions will require more research.
We will first consider two plants (\emph{henbane} and \emph{cannabis})
whose names are easily identifiable and where a visual comparison,
see Fig.~\ref{fig:Plants} can be considered as additional evidence
for the text-based identification. After this will report on some
preliminary attempts, i.e.~we are not attempting a botanical identification
of the plants \cite{plants} and should take into account that, even
in comparison to other medieval depictions, the drawings are far from
perfect.

\subsubsection{Henbane\label{sub:Henbane}}

\begin{figure}
\centering{}\includegraphics[height=5cm]{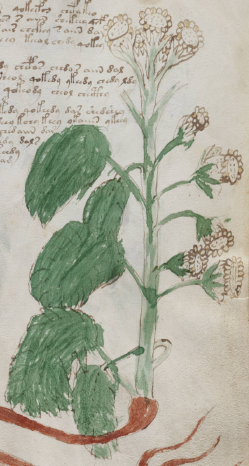}~~~~\includegraphics[height=5cm]{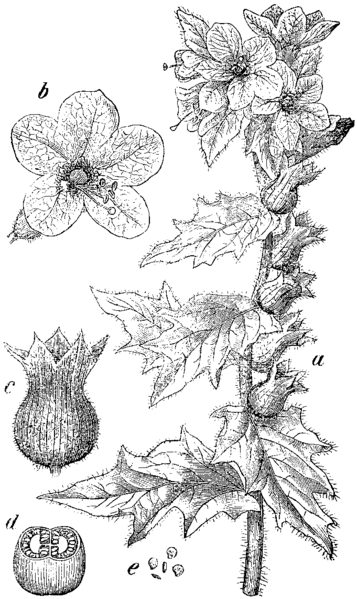}\caption{Henbane from VM f31r (left) and from Martin Cilen\v{s}ek's Na\v{s}e
\v{s}kodljive rastline, 1892 (wikimedia file: Nsr-slika-088), (right),
see Sect.~\ref{sub:Henbane}. While the leaf shape and the unilateral
position of the flowers roughly coincide, the shape of the flowers
is dissimilar. The VM may represent the ripened fruits of the plant.
\label{fig:Plants}}
\end{figure}

The first word of f31r is \emph{BccNcy} which can be transcribed as
\emph{bang}, see Tab.~\ref{tab:Letters-in-brackets}, which uniquely
translates \cite{MacKenzie} as \emph{henbane} (\emph{hyoscyamus niger}),
a poisonous plant of the nightshades family. The similarity of the
drawing on f31r and the plant henbane is illustrated by Fig.~\ref{fig:Plants}
and can be considered as additional evidence for the translation.

\subsubsection{Cannabis\label{sub:Cannabis}}

\begin{figure}

\begin{centering}
\includegraphics[height=5cm]{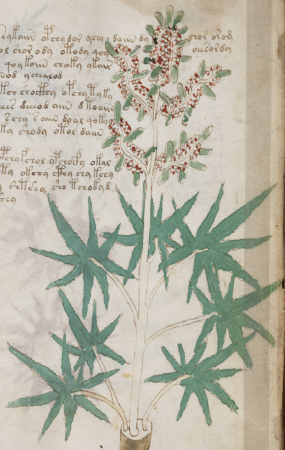}~~~~\includegraphics[height=5cm]{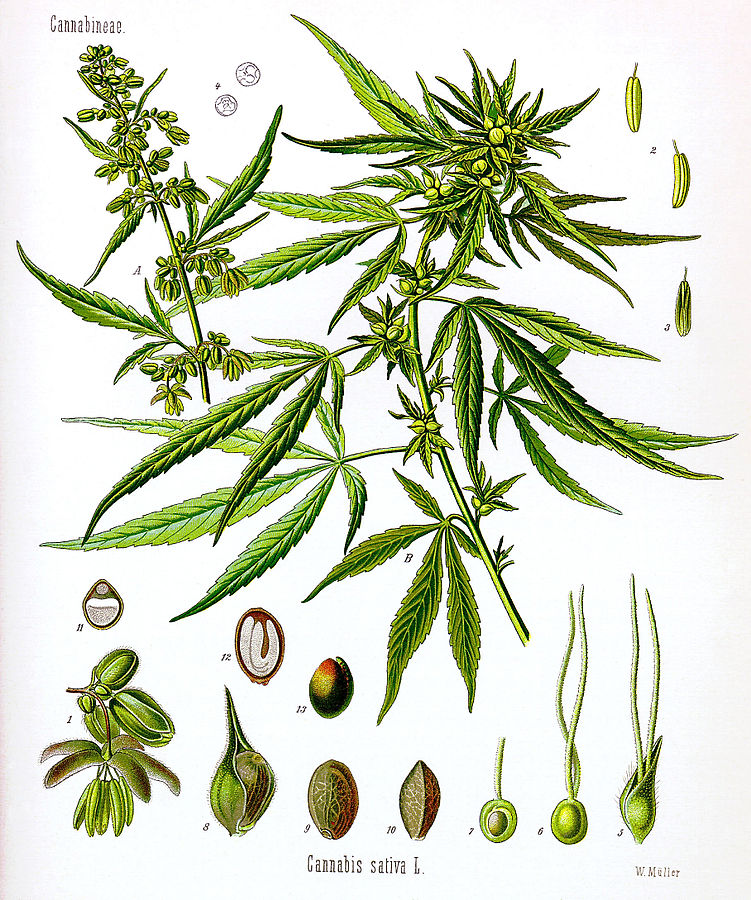}
\par\end{centering}

\caption{Cannabis.\label{fig:Cannabis.} from VM f16r (left) and from Franz
Eugen Köhler's Medizinal-Pflantzen, 1887. (wikimedia id: 1739269),
see Sect.~\ref{sub:Cannabis}.}
\end{figure}

The first word of f16r is \emph{\v{s}coN }which can be transcribed
as \emph{\v{s}\={a}n}, see Tab.~\ref{tab:Letters-in-brackets},
which uniquely translates \cite{MacKenzie} as hemp (\emph{cannabis}).
The similarity of the drawing on f16r and to the cannabis plant is
obvious: Leaves are neither clearly opposite nor alternating, they
are shown to consist of seven to nine finger-like leaflets. The spike-shaped
flowers are probably female and are riddled with elongated leaflets,
see Fig.~\ref{fig:Cannabis.}.

\subsubsection{Further observations on the plant pages}

Among the first words on the plant pages, we find often \emph{\v{s}P\={\i}g}
(sprout), \emph{d\={a}n }(grain), or \emph{d\={a}r} (tree) which
may be a general term or a component of a plant name that consists
of more than one word. E.g.~on f17r we find \emph{don }(\emph{d\={a}n})\emph{
}which here, however, may refer to buckwheat.

Folio 21r shows a plant similar to box (\emph{buxus}) or P\emph{ \v{s}im\v{s}\={a}r}.
The first word of the text is \emph{Sor} (\emph{\v{s}\={a}r}). Near
the end of the 7th line we find \emph{\v{s}om\v{s}or} (with the
middle \emph{m} and \emph{\v{s}} as an odd ligature).

The first word of Folio 24r can be read as \emph{al\={a}lag} which
would mean \emph{anemone }(\emph{anemone blanda} (?)), but the picture
does hardly match, although the anemone family has a wide variety
of leave shapes and numbers of petals. The must be said for f45v which
starts with the same word. A problem with this reading is also that
we otherwise ignored the \emph{waw} after the initial ``paragraph
marker'' \emph{P }while it would be part of the plant name here.

Chick-pea in P is \emph{nax\={o}d} which can be found in the beginning
of 5th line of 26v. While the leaves are may be plausible, the drawn
flowers are less typical, perhaps chick-peas are mentioned only for
comparison here and, therefore, not in the beginning of the text.

The drawing on f41v is identified as coriander (\emph{Coriandrum sativum})
\cite{Bax}, or \emph{gi\v{s}n\={\i}z} in P. Ref.~\cite{MacKenzie}
gives also the variant \emph{ki\v{s}n\={\i}z}. While the single
word that makes up the first line is unrelated to this P word, the
second line appears to give several variants, e.g.~the 2nd word contains
\emph{K\v{s}, }the third word reads\emph{ K\v{s}nd}.

\emph{Date palm} in P us \emph{mu$\gamma$} \cite{MacKenzie} which
can be found as the third word of the first line in f56r. The drawing
shows a plant with at least the base of the stem and the lower pair
of leaves reminiscent of a palm tree.

\subsubsection{Plant parts}

\begin{figure}
\centering{}\includegraphics[width=9cm]{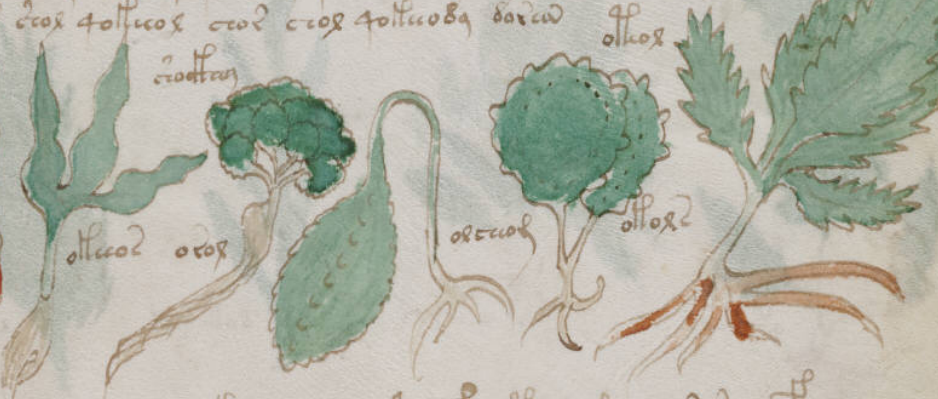}\caption{Shapes of plant parts (f100r)\label{fig:Shapes-of-plant}}
\end{figure}

Folio 100r gives an overview of shape types, most likely of leaves.
It contains six descriptive words for five pictures, see Fig.~\ref{fig:Shapes-of-plant}.
We therefore, consider the first word in the upper row (\emph{d\={o}sp\={\i}g},
i.e.~double spouted) is seen as the last word of the previous text.
The second word in the upper row \emph{Mht} could be \emph{mih} (false,
opposite) or \emph{mahist} (greatest). The four words in the second
row are less ambiguous. We have \emph{B\={\i}or} (\emph{bahr}, part)
for a leaf consisting of three parts, \emph{rot} (r\={o}d, river)
for a set of leaf stems that branch off like a river delta, \emph{t\v{s}hg},
which may relate to (\emph{ta\v{s}t}, bowl), and \emph{Botr} (\emph{BATR},
a heterogram for\emph{ pas} \cite{MacKenzie}, behind) for one leave
behind the other. This again is to be seen as evidence for our main
hypothesis rather than as an exercise in Pahlavi transcription.

\subsection{Lunar mansions}

In a similar way, it is possible to transcribe from the illustration
on f69v most of the 28 lunar mansions that are also listed in Bundahesh
II, 2 \cite{JustiBund}. Because of the short and repeated Pahlavi
names of the mansions, a unambiguous correspondence was possible for
only 20 of the mansions, such that we did not include it here. Interestingly,
the 1247 stone representation of the Suzhou star chart (1193) that
shows the related 28 Chinese constellations has a ``cartouche''
title beginning with the ideogram for \emph{sky} that can also be
seen in a corrupted and reverted form on f1r of the VM. This is not
implausible considering the continuous exchange between Persia and
China in historic times.

\subsection{Zoroastrian material}

The four words in the center of f67v2 are (with transcription) \emph{zoahd}
(\emph{zohr}), \emph{oBarao} (\emph{bahr}), \emph{zary} (\emph{z}\={o}\emph{r})
and \emph{natag} (\emph{nihadag}). The translation yields the words
\emph{sacrifice}, \emph{lot}, \emph{power}, \emph{foundation} \cite{MacKenzie}
that appear, given a Zoroastrian parentage, semantically related.
The words are grouped around a small square-shaped picture of a swirl-radiating
star which could represent a sacrificial fire.

App.~\ref{sec:Sample-text-(f1r)} includes a transcription of words
from the beginning of the third paragraph of f1r. This sample is included
not only to show that the Pahlavi transcription applies to the main
text, too, but also to demonstrate the difficulty of a translation
of the text, which has, however, been noted by all translators of
Pahlavi documents.

In the illustration on f77v, we find the words \emph{oBam yHat otBaNat
orShNat oMot dhNy oMotor} which can be transcribed as \emph{b\={\i}m
duxt wad-baxt r\={e}\v{s}inad m\={\i}h d\={e}n wiz\={a}r} and
is translated word-by-word as \emph{fear daughter unfortunately wounded:
false }(\emph{alternative?})\emph{ religion explanation} \cite{MacKenzie}.
This sample, nevertheless, suggests that the ``nudes'' pages (f75r
-- f84v) represent medieval medical content. While the representation
of nude bodies is rare in such contexts, similar scenes appear in
contemporary miniatures from Mughal India, where, however, an erotic
perspective is taken, which is not obvious in the VM.

\subsection{The colophon (f116v)}

Further evidence for the proposed transcription can be obtained from
the ``colophon'' (f116v). The last line of the short text contains
the words \emph{arar dccy} that are, in contrast to the seemingly
Latin script on this page, clearly readable. We propose the transcription
\emph{xwar day}, which would refer to the 11th day of the 10th month
of the Zoroastrian calendar \cite{MacKenzie}. The question whether
the $\cap\!\cap$ character before the lacuna at the end of this line
was originally the initial character of a year, cannot be answered
without further analysis of the velum. 

Based on the Pahlavi hypothesis presented here, it seems possible
to extract more information from the colophon. In the App. C, we present
an attempt to read the colophon, which, however, is largely speculative,
even if we assume that the Pahlavi hypotheses is true.

\section{Discussion }

We have not been presented more than a few words, which is mainly
due to the inherent difficulties in reading Pahlavi. Therefore, at
this point it is not clear, whether the VM contains also words of
a different idiom, such as the northwest Indian language Gujarati,
whose Parsi dialect contains many Iranian words due to the Zoroastrian
influence, although Gujarati does not itself identify as an Iranian
language. 

It is striking that the manuscript does not contain any obvious religious
symbolism (apart from the crucifix on f79v, which may well be a later
insertion) nor any other culturally identifiable elements. However,
the astronomical charts of the VM are related to the world of the
Zoroastrian culture in the middle East or South Asia. They do not
show any awareness of (earlier?) Arabic astronomy, but seem to follow
the cosmological view in the Bundahesh. 

Finally, we want to emphasise that we have no evidence that the VM
was produced in Persia (or perhaps even western India). It is also
possible that it originates from the regions near the Black Sea where
an exchange between Persia and the Italian cities of Genoa and Venice
took place around the presupposed time of the production of the VM.
Our opinion that the content of the VM is meaningful does not exclude
the possibility that it is still a ``hoax'', in the sense that it
was copied to be sold rather than read. In this process or by later
action, foliae with critical content may have been removed to further
obscure the origin of the manuscript. 

Although the proposed transcription is obviously tentative, it is
now possible to find many of the VM words in a Pahlavi dictionary
\cite{MacKenzie,Nyberg,JustiBund} using Table~\ref{tab:Letters-in-brackets},
which will give at least partial insight into the content of the VM.
We are also unable to provide a more precise phonemic account at this
stage, although some of the differences (e.g.~between V \emph{d}
and P \emph{t}) may allow for such discussions. It will require a
substantial effort to provide a complete translation of the VM, as
it seems unlikely that large parts of the text have been passed down
also from other sources, i.e.~the VM does not appear to be identical
to any of the better known Zoroastrism-related scriptures or commentaries,
so its content may as well have an interest on its own.

\subsection*{Acknowledgement }

The author acknowledges the use of the high-resolution scans made
available by Jason Davies' Voynich Voyager \cite{Voyager}. We are
also grateful to two anonymous reviewers for their comments on an
earlier version of this manuscript. This earlier version is available
as a working paper from the PURE repository (University of Edinburgh)
since 1st of August, 2017.

This is the second version of the paper. It differs from the first version
by the deletion of some but not all superfluous text and by a 
reference~\cite{Peterson2005} to an earlier mentioning of the 
Pahlavi hypothesis.

\bibliographystyle{abbrv}
\bibliography{voy}

\begin{thebibliography}{10}

\bibitem{Amancia}
D.~R. Amancio, E.~G. Altmann, D.~Rybski, O.~N. Oliveira~Jr, and L.~d.~F. Costa.
\newblock Probing the statistical properties of unknown texts: {A}pplication to
  the {V}oynich manuscript.
\newblock {\em PLoS ONE}, 8(7):00--00, 2013.

\bibitem{Andreas}
F.~C. Andreas and K.~Barr.
\newblock Bruchstücke einer {P}ehlevi-\"{U}bersetzung der {P}salmen.
\newblock In {\em Sitzungsberichte}. Preu{\ss}ische Akademie der
  Wissenschaften, Berlin, 1933.

\bibitem{Bax}
S.~Bax.
\newblock A proposed partial decoding of the {V}oynich script, 2014.

\bibitem{Buehler}
G.~B\"uhler.
\newblock {\em On the Origin of the Indian Brahma Alphabet}.
\newblock K. J. Trübner, 1898.

\bibitem{Daniels}
P.~Daniels and W.~Bright.
\newblock {\em The world's writing systems}.
\newblock Oxford University Press, New York, Oxford, 1996.

\bibitem{Voyager}
J.~Davies.
\newblock Voynich manuscript voyager, 2013.
\newblock http://www.jasondavies.com/voynich/.

\bibitem{EVA}
E{V}{A}.
\newblock see http://www.voynich.nu/extra/eva.html.

\bibitem{Faulmann}
C.~Faulmann.
\newblock {\em Das {B}uch der {S}chrift enthaltend die {S}chriftzeichen und
  {A}lphabete aller {Z}eiten und aller {V}\"olker des {E}rdkreises}.
\newblock Verlag der Kaiserlich-k\"oniglichen Hof- und Staatsdruckerei, Wien,
  2nd edition, 1880.

\bibitem{GeigerandKuhn2002}
W.~Geiger and E.~Kuhn, editors.
\newblock {\em Grundriss der iranischen Philologie}, volume I.1.
\newblock Adamant, Boston, 2002.

\bibitem{Jaskiewicz}
G.~Jaskiewicz.
\newblock Analysis of letter frequency distribution in the {V}oynich
  manuscript.
\newblock In M.~S. et~al., editor, {\em Concurrency, Specification and
  Programming. Proceedings of the international workshop CS\&P}, page 250,
  Pullusk, Poland, 2011.

\bibitem{JustiBund}
F.~Justi.
\newblock {\em Der Bundehesh}.
\newblock F. C. W. Vogel, Leipzig, 1868.
\newblock (Note that pages in the Pahlavi part are in incorrect order).

\bibitem{ZCalendar}
A.~Lali.
\newblock Zoroastrian calender at {ZCS}erv, 2005-2017.
\newblock http://www.zcserv.com/ \nobreak{calendar}/.

\bibitem{MacKenzie}
D.~N. MacKenzie.
\newblock {\em A concise Pahlavi dictionary}.
\newblock Oxford University Press, London, 1986.

\bibitem{Montemurro}
M.~Montemurro and D.~Zanette.
\newblock Keywords and co-occurrence patterns in the {V}oynich manuscript: {A}n
  information-theoretic analysis.
\newblock {\em PLoS ONE}, 8(6):0066344, 2013.

\bibitem{Mueller}
F.~M. M\"uller, editor.
\newblock {\em The sacred books of the {E}ast}, London, 1879-1910. Macmillan
  and Co.

\bibitem{Nyberg}
H.~S. Nyberg.
\newblock {\em A Manual of Pahlavi, Pt. 2}.
\newblock Otto Harrassowitz, Wiesbaden, 1974.

\bibitem{Paulesu}
E.~Paulesu, N.~Brunswick, and F.~Paganelli.
\newblock Cross-cultural differences in unimpaired and dyslexic reading:
  Behavioral and functional anatomical observations in readers of regular and
  irregular orthographies.
\newblock In N.~Brunswick, S.~McDougall, and P.~de~Mornay~Davies, editors, {\em
  Reading and Dyslexia in Different Orthographies}, chapter~12. Psychology
  Press, 2010.

\bibitem{Peterson2005}
J.~H. Peterson.
\newblock {\em VMs: RE: flowers \& stars: Zoroastrian stuff, Pahlavi}.
\newblock vms-list, 23 Feb. 2005.

\bibitem{Reddy}
S.~Reddy and K.~Knight.
\newblock What we know about the {V}oynich manuscript.
\newblock In {\em Proc. of the 5th ACL-HLT Workshop on Language Technology for
  Cultural Heritage, Social Sciences, and Humanities}, pages 78--86, Madison,
  WI, 2011. Omnipress, Inc.

\bibitem{Rugg2004}
G.~Rugg.
\newblock An elegant hoax? {A} possible solution to the {V}oynich {M}anuscript.
\newblock {\em Cryptologia}, 28(1):31–46, 2004.

\bibitem{Schninner2007}
A.~Schinner.
\newblock The {V}oynich {M}anuscript: {E}vidence of the hoax hypothesis.
\newblock {\em Cryptologia}, 31(2):95–107, 2007.

\bibitem{GabrIranica}
M.~Shaki.
\newblock Entry: Gabr.
\newblock {\em Encyclopedia Iranica}, X(3):239--240, 2000, 2012.

\bibitem{plants}
E.~Sherwood and E.~Sherwood.
\newblock The {V}oynich botanical plants, 2008, accessed on 5/9/2017.
\newblock http://edithsherwood.com/voynich\_botanical\_plants.

\bibitem{WestBundahesh}
E.~W. West.
\newblock {\em The Bundahishn (``Creation''), or Knowledge from the Zand},
  volume~5 of {\em Sacred Books of the East}.
\newblock Oxford University Press, London, 1897.

\end{thebibliography}

\newpage

\appendix

\section{Zodiac pages (f70v1-f73v)\label{sec:Zodiac-signs-(f70v1-f73v)}}

All descriptions were found within the V script around the margin
(for f70v2, within the margin) of the central image that shows a depiction
of the zodiacal sign.

\subsection*{$\star$ Notes}

\Zodiac{10} The centre image shows Aries in repetition of f70v1,
but also the text does not show much evidence for the interpretation
as Capricorn. The first letters of the V constellation are ignored,
so the transcription is questionable.

\Zodiac{11} The centre image shows Taurus in repetition of f71v,
but also the text does not show much evidence for the interpretation
as Aquarius. The first letters of the V month and V constellation
are ignored, so the transcription is questionable. Alternative spelling:
A\v s{}wahi\v s{}t \includegraphics[height=0.6cm]{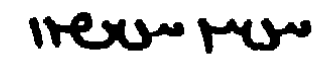}.

\Zodiac{8} Alternative spelling: \includegraphics[height=0.7cm]{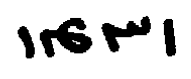}
Wahman.

\Zodiac{9} Alternative spelling: \includegraphics[height=0.7cm]{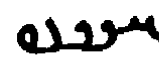}\includegraphics[height=0.6cm]{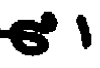}
nem$\bar{\mbox{a}}$sp.

\newpage\thispagestyle{empty}

\includegraphics[width=1\linewidth]{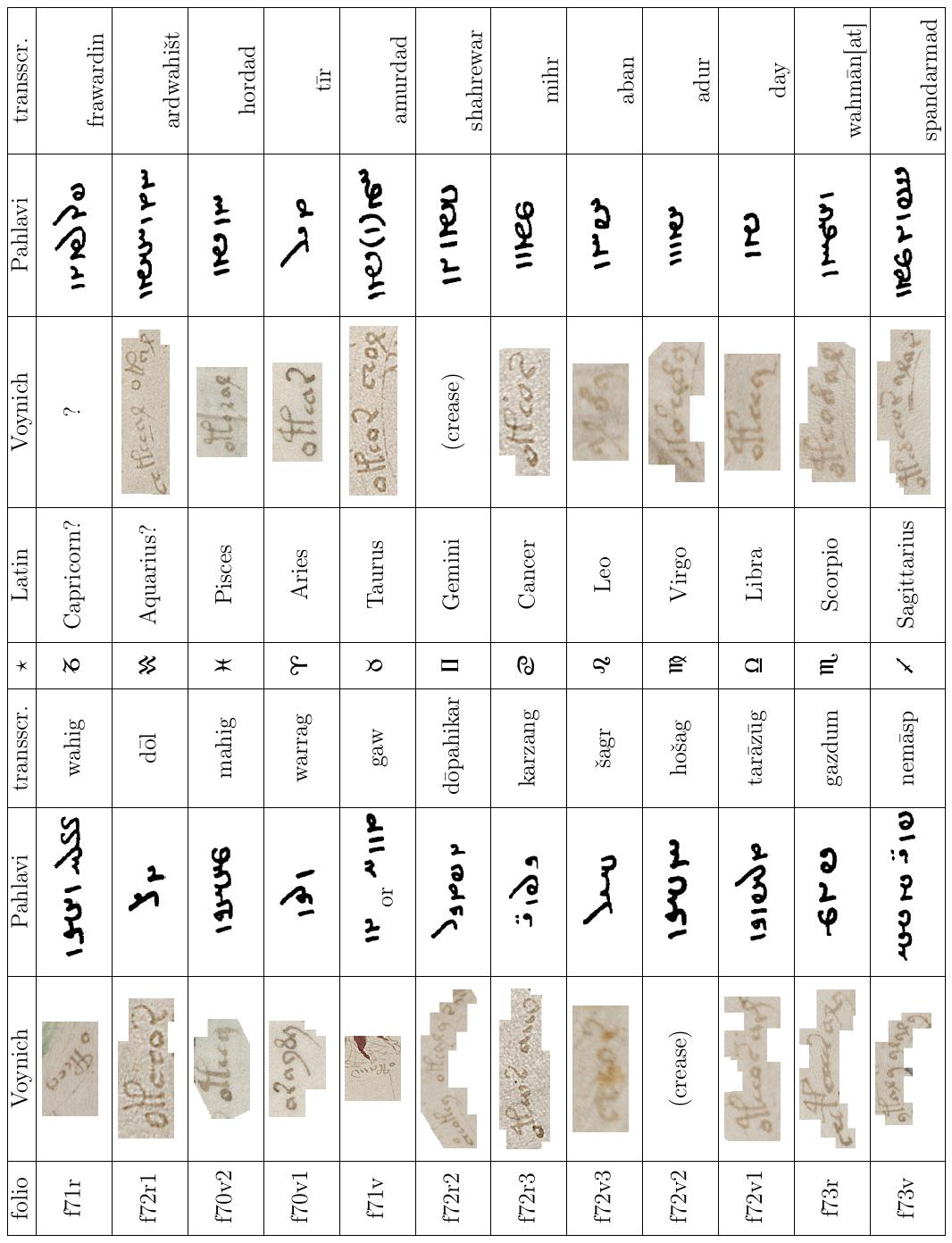}

\section{First folio text (f1r)\label{sec:Sample-text-(f1r)}}

Passage from the beginning of the third paragraph of f1r. Not all
translations from \cite{MacKenzie} are shown. Our transliteration
shows several inconsistencies, which may be due to the complexity
and development of the Pahlavi language and will require further analysis.
E.g. V \emph{otr} retains P \emph{t}, while V \emph{dody} uses \emph{d}
for P \emph{t} in accordance with the transliteration \cite{MacKenzie}.
Final \emph{o}, as in P for \emph{dody}, is often ignored as an \emph{otiose
stroke} \cite{MacKenzie}, see also the final character of \emph{Spandarmad}.
In VM, more often leading \emph{o} are otiose, e.g. in \emph{Spandarmad}
after the line break, while in V \emph{otr} the leading \emph{o} is
considered as part of the word. 

\includegraphics[width=1\linewidth]{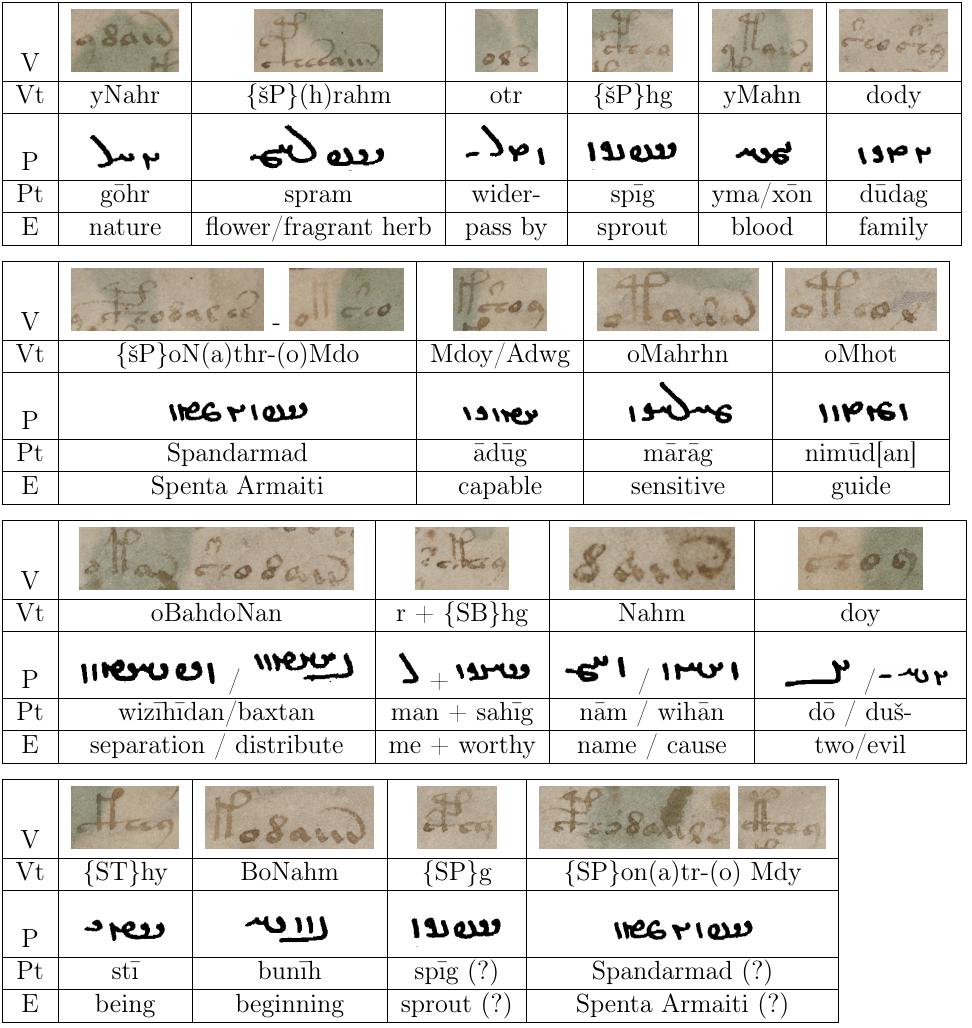}

\subsection*{Notes}

Curly brackets enclose \{ligatures\}. Square brackets indicate {[}inserted
characters{]}. Round brackets indicate (ignored characters). A hyphen
stands for a line break. Small strokes appearing in V either as \emph{c}
or \emph{\i} are transcribed here as \emph{h}, i.e.~are considered
to indicate a lengthening of the nearest vowel.

To explain the ignored V \emph{a} in \emph{Spandarmad}, the P \emph{d}
could be considered as a contraction of V \emph{a} and V \emph{t}.

The chapter on \emph{The Nature Of Plants} (Bundahesh, Ch.~XXVII)
mentions \emph{Spandarmad} \cite{WestBundahesh}

\section{The colophon of the Voynich manuscript}

\subsection{Introduction}

The last page (f116v) of the Voynich Manuscript (VM) can be assumed
to show a colophon, i.e. an addendum that occurs frequently usually
on the final page of medieval manuscripts and early modern prints,
which usually contains information on the author, production, provenance
etc. Following the hypothesis of the main text that the VM is written
in a Pahlavi-like script, we present a translation of most of the
colophon text. We can identify a place of origin and a date, but not
the year in which the manuscript was written. Also we believe to be
able to identify the scribe's name which may refer to a female writer
from a medieval Zoroastrian community possible in the city of Trebisonta,
the a gateway between Persia, Byzantium and early renaissance Italy. 

\begin{figure}[h]
\begin{centering}
\includegraphics[width=1\linewidth]{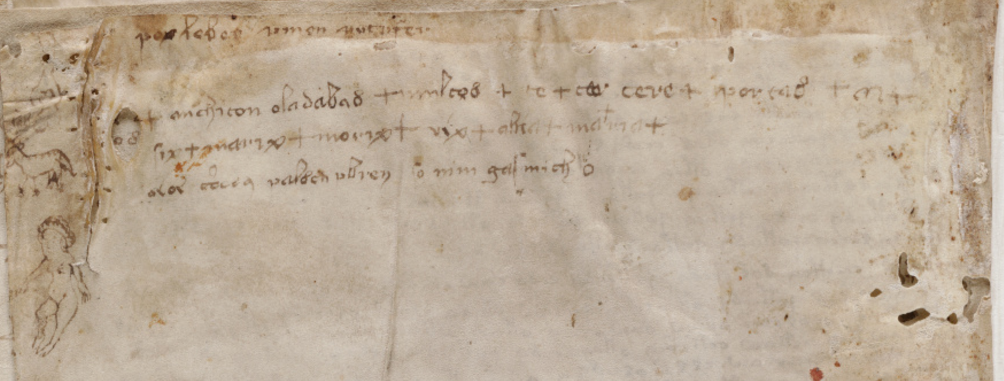}
\par\end{centering}

\caption{Full view of the relevant part of the colophon page (f116v).\label{fig:Full-view-of}}
\end{figure}

The interpretation is in large part speculative and in need of further
research, but is included here in order to stimulate further study
as well as to provide in turn additional evidence for the Pahlavi
hypothesis. 

Below we will consider the last page of the VM line-by-line in some
detail based on the Pahlavi hypothesis. This will enable us to draw
some conclusions on the context in which the content of the VM may
have originated.

\subsection{Background}

\subsubsection{Pahlavi hypothesis}

The claim that the VM is written in natural language rests on the
observations that most of the Voynich (V) characters have counterparts
in the middle-Iranian Pahlavi script that was used in medieval Zoroastrian
scriptures, commentaries, and a few other texts \cite{Andreas}. Some
of the V characters characters are upside-down versions of their Pahlavi
counterparts, which may be an effect of different writing directions.
Other V letters are related in another obvious way or can be explained
as ligatures. Finally, two letters are added, but are easily identifiable
from the vocabulary. In principle, it is thus possible to translate
Voynich words using a Pahlavi dictionary such as Ref.~\cite{MacKenzie}.
Although this process is in many cases successful, is is not always
straightforward. 

The colophon of the Voynich manuscript contains only a few Voynichese
letters. Most of the characters resemble Latin letters, but the awkwardness
of their shapes contrasts strikingly with the fluency of the proper
Voynichese letters. This indicates the possibility that the colophon
text was written by a scribe not well acquainted with in Latin letters.
Also the impossibility of identifying any other of the language that
typically uses Latin script, justifies the attempt to identify the
Latin letters as transcription from Voynichese.

\subsubsection{Colophon }

A colophon in medieval European manuscripts usually starts with the
\emph{explicit} that contains the Latin phrase \emph{explicit liber}
(the book is ``spread out'', i.e.~finished), although since early
modern times the words \emph{colophon} and \emph{explicit} are used
interchangeably. After the explicit, the colophon may give information
about content, author, place, date, producer, commissioner, the publication
process etc.

\begin{figure}[H]
\begin{centering}
\includegraphics[scale=0.33]{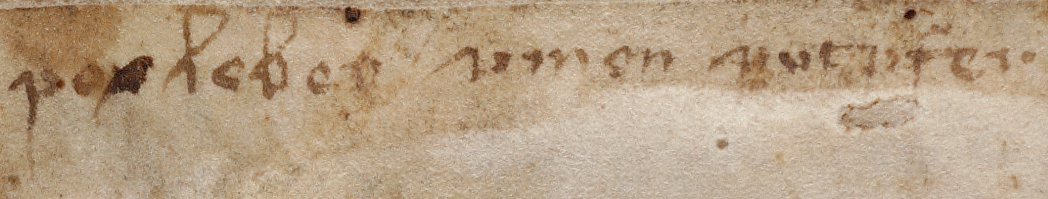}
\par\end{centering}

\caption{The ``title'' of the colophon page.\label{fig:The-title-of}}
\end{figure}

\begin{figure}[H]
\begin{centering}
\includegraphics[scale=0.5]{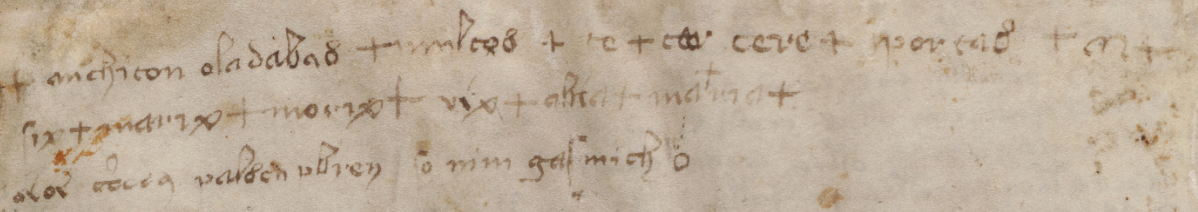}
\par\end{centering}

\caption{The colophon.}
\end{figure}

\subsection{First line}

\subsubsection{Explicit}

The beginning of colophons often reads \emph{explicit liber}. We
will try to identify a similar Pahlavi expression in the beginning
of the first line. 

With this bias, we propose to read the first word as\emph{ m\={a}day\={a}n}
(\emph{book}). The letters \emph{a} in the manuscript are seen to
indicate lengthening of the vowel that is not written in abjad alphabets.
The first \emph{a} touches the following letter and the second one
is corrected \={a} by an overlapping \emph{o}. The \emph{i} is meant
to represent Pahlavi \emph{y}. 

The word ``be finished'' in Pahlavi is \emph{frazaftan} (alternatively\emph{
han\v{\j}aftan}), which we can read in the second word of the line.
We need to assume the leading \emph{o} is not a word separator but
stands as a \emph{waw} for the \emph{f} sound (usually \emph{f} is
expressed by \emph{p} in Pahlavi). Reading the second letter as \emph{r}
is consistent with all occurrences of this character on this page.
Another critical assumption is the interpretation of the sixth letter
as a ligature of f and t or by an omission of one of the two letter.
Also the other letters are unambiguous including the final figure-8
character that always represents an $n$. We should note again, that
an interpretation of the first two words without being biased by the
expectation of the \emph{explicit}, would be very difficult.

\begin{figure}[H]
\begin{centering}
\includegraphics[scale=0.45]{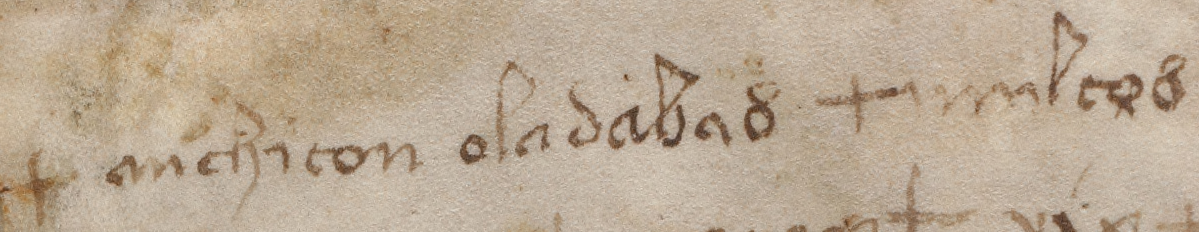}
\par\end{centering}

\caption{First part of the first line. This figures as well as the following
ones have the same scale.\label{fig:anchiron}}
\end{figure}

The last word in Fig.~\ref{fig:anchiron} is read as Pahlavi \emph{m\={a}rdan}
(spelt with \emph{t} \cite{MacKenzie}) for \emph{perceive}, \emph{notice}
or \emph{feel}. Also here a correction of a (the penultimate) letter
is seen. It may emphasise the fact stated by the first to words or
could, in the sense of \emph{done at} relate to the place name that
is seen to follow in this line, see Fig.~\ref{fig:trabzon}.

\subsubsection{Trabzon}

Today's Trabzon, was the antique town Trapezos ($T\!\rho\alpha\pi\varepsilon\zeta o\tilde{\upsilon}\varsigma$)
on the south-eastern shore of the Black Sea. It had an important role
as a trade gateway to Persia and was regularly called at by Venetian
trading ships during 13th and 14th centuries. As the capital of the
Empire of Trebisonta is was a melting pot of religions. In this way
it would be a plausible location for a Zoroastrian book to be transferred
from Persia to Europe.

\begin{figure}[H]
\begin{centering}
\includegraphics[scale=0.45]{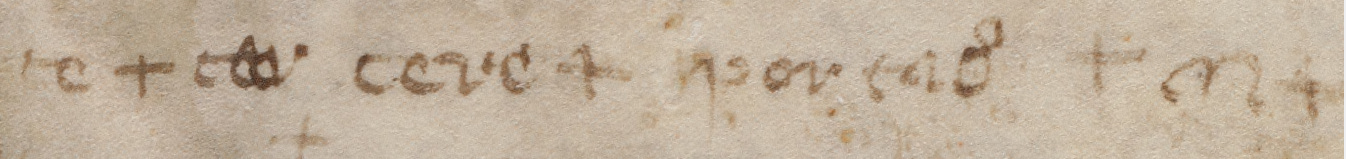}
\par\end{centering}

\caption{Second part of the first line.\label{fig:trabzon}}
\end{figure}

In Fig.~\ref{fig:trabzon}, we note that first two characters (with
a $+$ sign between them) appear as unsuccessful attempts to construct
a ligature that does not exist in Pahlavi. The combination \emph{\v{s}r}
does not occur in initial position \cite{MacKenzie}, where ligatures
are mostly used in other parts of the VM. Only at the third attempt,
the \emph{c}-shape is correctly placed between the legs of the $\pi$-shape
that is usually expressing the sound \emph{\v{s}}, but appears here to represent
\emph{t}. Although there is evidence elsewhere for this corruption,
it is clearly a weak point of the interpretation. Also the split of
the word into the parts \emph{treb} and \emph{isonta} casts doubts
on the identification. The final \emph{n} (figure \emph{8}-shape)
is less critical as it can be seen as a locative ending. We, nevertheless,
propose Trebisonta as the putative location of the production of the
VM. 

For the last letter of the first line, \emph{M}, refer to Sect.~\ref{sec:Illegible-characters-on}

\subsection{Second line}

\begin{figure}[H]
\begin{centering}
\includegraphics[scale=0.55]{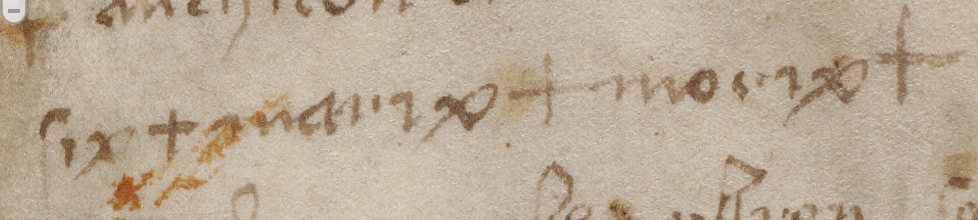}
\par\end{centering}

\caption{First part of the second line.\label{fig:First-part-of}}
\end{figure}

\begin{figure}[H]
\begin{centering}
\includegraphics[scale=0.5]{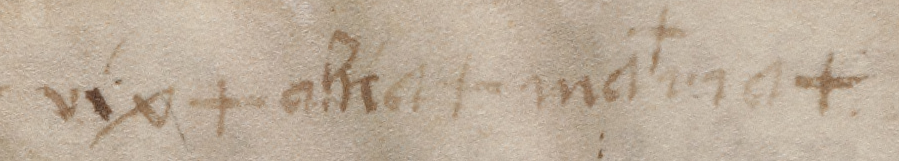}
\par\end{centering}

\caption{Second part of the second line.\label{fig:Second-part-of}}
\end{figure}

If we take the beginning of the second line (Figs.~\ref{fig:First-part-of}
and \ref{fig:Second-part-of}) as a direct continuation of the first
one (see, however, Sect.~\ref{sec:Illegible-characters-on}) and
identify the first letter as an \emph{r} (compare Sect.~\ref{sub:Finis}),
which is also used to denote the number $20$ \cite{MacKenzie}. It
may not seem straightforward to explain why the Pahlavi numeral is
followed by a Roman \emph{IX} (Pahlavi for number \emph{9} would be
\emph{333}), but it is not fully unexpected considering the organisation
of the Pahlavi tens in steps of 20. In combination with the \emph{M}
in the preceding line leads to an year 1029 which can refer (based
on the date in Sect.~\ref{sub:Date}) to the Christian (11th of November
1029), the Muslim (9th of September 1620) or the Zoroastrian era (19th
of July, 18th of August, or 26th of December in 1660, resp., for the
Kadmi, Shenshay or Fasli calendars \cite{ZCalendar}). However, from
the dating of the velum to 14th century, neither of these dates appears
likely. One possibility is to use the velum date to justify a lost
Roman \emph{CD} after the \emph{M} at the end of the first line, such
that a date of 1429 is implied, which is, however, highly speculative
and contradicts the use of the Zoroastrian calendar for the month
(Sect.~\ref{sub:Date}). Whether or not the space after the \emph{M}
contains indeed the minuscule letters \emph{cd} cannot be decided
from the available scans of the VM.

In the remainder of the line, we can identify in this line three attempts
to write the word \emph{m\={a}h} meaning \emph{moon }or\emph{ month.
}It consists of the letter \emph{m }followed by\emph{ a }for the lengthening
of the vowel and three strokes representing \emph{h}. In the second
occurrence of the word instead of \emph{a} the letter $o$ is written.
The first two occurrences precede what appears to be the Roman numeral
\emph{X}. The third occurrence of \emph{m\={a}h} follows a word with
the possible spelling\emph{ abha} (Fig.~\ref{fig:Second-part-of}).
The reading is not clear, but the word may be Latin for the Zoroastrian
day name \emph{xwar} that occurs also in the third line, see Sect.~\ref{sub:Date}.

\subsection{Third line}

\subsubsection{Date\label{sub:Date}}

\begin{figure}[H]
\begin{centering}
\includegraphics[scale=0.5]{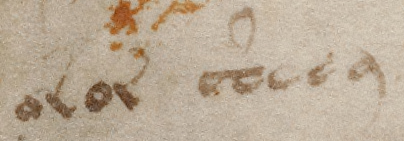}
\par\end{centering}

\caption{First part of the third line.\label{fig:aror}}
\end{figure}

The last line of the short text starts with the words \emph{aror dccy,
}see Fig.\emph{~\ref{fig:aror}} which are, in contrast to most of
the awkward Latin script on this page, unambiguously identifiable:
The letters \emph{a, o and r} can express the same Latin phonemes.
The combination \emph{cc} is a single letter which can refer to \emph{s}
or \={\i} and the last character is the ambiguous \emph{d}-\emph{g}-\emph{y}
letter mentioned in the introduction. The remaining letter \t{$\pi$}
functions as a \emph{d}. The Pahlavi correspondence is nevertheless
more complicated, but suggests unambiguously the irregular transcription
\emph{xwar day}, which refers to the 11th day of the 10th month of
the Zoroastrian calendar \cite[p.~142]{MacKenzie}. After the mediocre
attempts to give the date in Latin script in the previous line, it
seemed necessary to return to the more familiar and less ambiguous
Voynichese expression.

\subsubsection{Name}

\begin{figure}[H]
\begin{centering}
\includegraphics[scale=0.5]{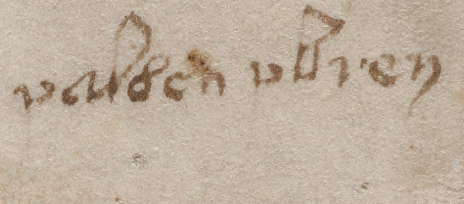}
\par\end{centering}

\caption{Second part of the third line. Phrase interpreted as the name of the
author: G\emph{\={o}}lnar Gabr\={\i}.\label{fig:Golnar-Gabr}}
\end{figure}

Observing the descenders in the two initials in Fig.~\ref{fig:Golnar-Gabr},
we read two first letters of the words as \emph{g,} as in the ``title''
of the folio f116v. The forth letter of the first word is, as in Figs.~\ref{fig:anchiron}
and \ref{fig:trabzon}, an \emph{n}, and the following letter an \emph{r},
see Fig.~\ref{fig:aror}. Considering the remaining characters as
Latin letters, we can identify the string \emph{Galnr Gbrey}. The
Persian name \emph{G\={o}lnar} refers to the flower of a pomegranate
tree, and is used since medieval times, as obvious from its prevalence
among the Parsis in India. Since Pahlavi uses essentially an abjad
alphabet, it seems natural that the vowel between the Voynichese letters
is omitted. Likewise, the second letter (\emph{a}) does not stand
for the vowel \emph{o}, but expresses the lengthening of the sound. 

\emph{Gbrey} is a Pahlavi from of \emph{Gabr }or \emph{Gabr\={\i}},
a term that was used for non-Muslim people in Iran. It seems to have
been applied mainly to members of the Zoroastrian faith \cite{GabrIranica}.
Considering that, when the VM was written, it would not have had the
later pejorative meaning, it could be have well been used as a byname,
and in fact has survived in several variants as a surname. 

\emph{G\={o}lnar Gabr\={\i}} is likely to a female name, although
also unisex names with the compound \emph{G\={o}l} (rose) exist.
If the scribe was indeed one of the rare female authors or writers
from that time, then the preservation of the VM is indeed very interesting,
even if we do not yet have much insight into the actual content of
the manuscript.

\subsubsection{Finis\label{sub:Finis}}

\begin{figure}[H]
\begin{centering}
\includegraphics[scale=0.5]{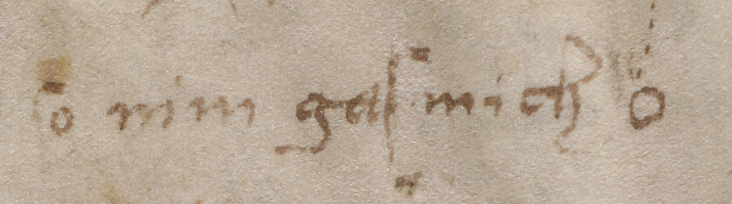}
\par\end{centering}

\caption{Final phrase of the colophon.\label{fig:So-nim}}
\end{figure}

As the least phrase of the colophon is particularly obscure, it is
hard to resist reading the last words of the colophon, Fig.~\ref{fig:So-nim},
as the German phrase \emph{So nimm gar mich.} (\emph{Thus take even
me}), which appears to be out off context. However, the identification
of the first letter as an descending \emph{s} and of the similar letter
in the middle as an \emph{r} is not entirely inconsistent within the
Pahlavi hypothesis, although we would expect the writer to return
rather to the Voynich \emph{r} in case of ambiguity.

A slightly more likely alternative reading can be found by considering
the phrase in Fig.~\ref{fig:So-nim} again as Pahlavi written in
Latin letters. In this case we would ignore the first part 

If we read the first letter as \emph{b}, then we can identify the
first word as\emph{ band} which refers to \emph{bastan} and is occasionally
written as \emph{bn} \cite{MacKenzie}, noting that the letter \emph{waw}
(\emph{o}) and \emph{nun} (\emph{n}) are of identical shape in Pahlavi.
The word \emph{bastan} means \emph{tie} or \emph{bind}. Other options
of starting with \emph{b }would require the presence of a third letter. 

We prefer to read the first letter as \emph{r} (as in the third word
in Fig.~\ref{fig:So-nim}), for which MacKenzie \cite{MacKenzie}
suggests the transliteration \emph{raw} which then refers to \emph{raftan}
meaning \emph{go} or \emph{move}. The following words can be transcribed
\cite{MacKenzie} as \emph{nim}{[}\emph{ay}{]} (\emph{nimudan}) meaning
\emph{guard, g\={o}r} for \emph{nature} (or \emph{jewel}) and \emph{mizd}
which can mean\emph{ reward}. 

Although the rough translation of the phrase as ``Go, guard nature's
reward!''~appears utterly anachronistic, it would be within context
and could be considered as a final message of the author to posterity.
Nevertheless, as many of our conclusions are rather speculative, this
translation is even more so.

\subsection{Drawings}

\begin{figure}[h]
\begin{centering}
\includegraphics[scale=0.5]{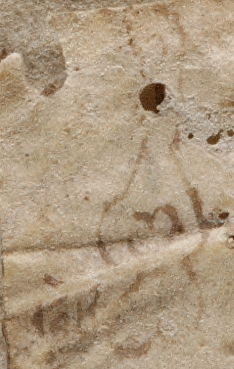}~~~\includegraphics[scale=0.5]{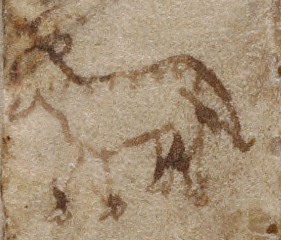}~~~\includegraphics[scale=0.5]{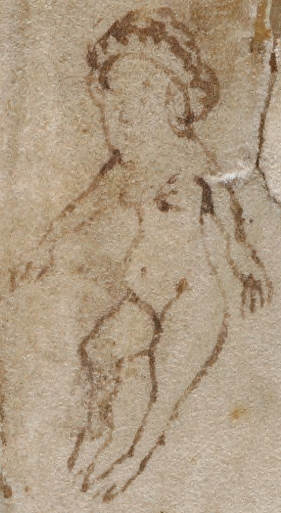}
\par\end{centering}

\caption{Drawings on left margin of f116v. The thee images (left to right)
are positioned on top of each other and separated from the main text
by a repaired tear in the velum. \label{fig:Drawings}}
\end{figure}

The velum of the last page has tear that apparently has been mended
before the use of the page. The scribe used the large part right of
the tear for text of the colophon, and decorated the margin left of
the tear by a few small drawings, see Fig.~\ref{fig:Drawings}.

The top picture (Fig.~\ref{fig:Drawings} left) has a conspicuous
likeness to a chicken corpse. The middle picture represents a billy
goat or a similar animal. The bottom pictures is a female nude in
the style of the figures in the ``nudes'' pages (f75r -- f84v).
It would be strange to assume that it represents the author. 

Although we are unable to give a interpretation of the pictures here,
we note the letters in the top figure (Fig.~\ref{fig:Drawings} left).
Although the first letter as a similarity with \emph{F}, we should
stay with the earlier reading of the character as \emph{r}, which
is followed by \emph{a} (or \emph{o}) and \emph{n} (figure-$8$ shape).
The word \emph{r\={a}n} means \emph{fight} \cite{MacKenzie}, but
our confidence in this reading is low.

\subsection{Illegible characters on right margin\label{sec:Illegible-characters-on}}

\begin{figure}[H]
\begin{centering}
\includegraphics[scale=0.5]{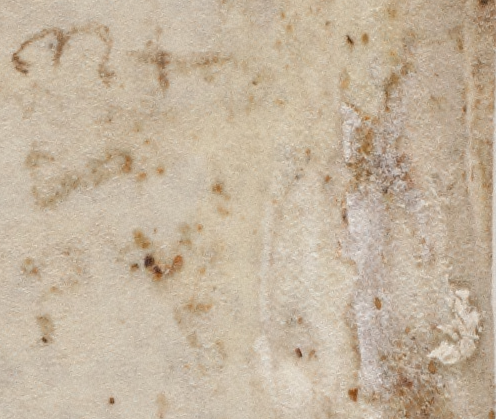}
\par\end{centering}

\caption{Ending of the first line of the colophon and lacuna due to abrasion
at the right margin of f116v.\label{fig:Ending-of-the}}
\end{figure}

One of the most important information to obtain from a colophon would
be the year of the production of the document. We have touched upon
this question above, but are unable to give a definite answer. Fig.~\ref{fig:Ending-of-the}
shows the ending of the first line of the colophon and the right margin
of the page next to the colophon. It is possible that characters after
the letter \emph{M} are lost, although they may become visible in
a multi-spectral analysis of the velum. 

In the left middle and lower part of Fig.~\ref{fig:Ending-of-the}
a few blurred characters (such as a question mark) can be seen, but
due to the difference in the strokes and unrelatedness to the main
part of the colophon, they should be considered as later additions.

\subsection{Conclusion}

All of the presented results from our reading of the colophon are
to a larger or lesser extent speculative, but at least we should admit
that the combined evidence provided here shows that the text on f116v
indeed represents a colophon. Further analysis of the text as well
as of the material document may lead to more reliable information
about the manuscript.

Colophons occur already on Ancient Near East clay tablets, but in
the case of the VM, the addition of a colophon appears as an ineffectual
attempt to adopt a Western custom, which may have be seen as suboptimal
when taken, but we can now appreciate it as potentially very useful.
\end{document}